# Local Markov Property for Models Satisfying Composition Axiom

**Changsung Kang**
Department of Computer Science
Iowa State University
Ames, IA 50011
*cskang@iastate.edu*

**Jin Tian**
Department of Computer Science
Iowa State University
Ames, IA 50011
*jtian@cs.iastate.edu*

## Abstract

The local Markov condition for a DAG to be an independence map of a probability distribution is well known. For DAGs with latent variables, represented as bi-directed edges in the graph, the local Markov property may invoke exponential number of conditional independencies. This paper shows that the number of conditional independence relations required may be reduced if the probability distributions satisfy the composition axiom. In certain types of graphs, only linear number of conditional independencies are required. The result has applications in testing linear structural equation models with correlated errors.

## 1 Introduction

The use of graphical models for encoding distributional and causal assumptions is now fairly standard (see, for example, [Pearl, 1988, Spirtes *et al.*, 1993, Pearl, 2000]). The most common such representation involves a directed acyclic graph (DAG), called a Bayesian network, over a set of variables. The statistical information encoded in a Bayesian network is completely captured by conditional independence relationships among the variables. The set of conditional independence relationships encoded in a DAG can be read by d-separation criterion, which provide the global Markov property for DAGs. A local Markov property specifies a much smaller set of conditional independencies that will imply (using the laws of probability, typically semi-graphoid axioms) all other conditional independencies which hold under the global Markov property. A well-known local Markov property for DAGs is that each variable is conditionally independent of its non-descendants given its parents.

When some variables in a DAG model are not observed, called *latent* or *hidden* variables, DAGs with bi-directed edges ($\leftrightarrow$) have been used to represent the conditional independence relations among observed variables [Pearl, 2000, Richardson and Spirtes, 2002]. DAGs with bi-directed edges have also been used to represent linear structural equation models (SEMs) with correlated errors, called path diagrams [Wright, 1934]. A DAG with bi-directed edges is called an acyclic directed mixed graph (ADMG) in [Richardson, 2003]. A natural extension of the d-separation criterion, called m-separation (see Section 2.1) can be applied to ADMGs which provides the global Markov property for ADMGs [Spirtes *et al.*, 1998, Koster, 1999, Richardson, 2003]. A local Markov property for ADMGs is given in [Richardson, 2003], which, in the worst case, may invoke an exponential number of conditional independence relations, a sharp difference with the local Markov property for DAGs, where only one conditional independence relation is associated with each vertex.

In this paper, we seek to improve the local Markov property for ADMGs given in [Richardson, 2003] in the situation that the probability distributions also satisfy the composition axiom. The intended application is in linear SEMs, which are widely used in the social sciences and economics [Duncan, 1975, Bollen, 1989]. In a linear SEM, variables are typically assumed to have normal distribution, and conditional independence relations will correspond to zero partial correlations, namely, a partial correlation $\rho_{xy.Z} = 0$ if and only if $x$ is independent of $y$ given $Z$. An important task in SEM applications is to test the model against data. The conventional method involves fitting the covariance matrix, while recently an alternative approach has been proposed which involves testing for the vanishing partial correlations [Spirtes *et al.*, 1998, Pearl, 2000]. The advantages of using such local tests instead of the traditional global fitting tests are discussed in [Pearl, 2000]. The path diagrams for linear SEMs without correlated errors are DAGs, and based on the local Markov property for DAGs, only one vanishing partial correlation test is needed for each variable [Pearl and Meshkat, 1999]. On the other hand, the path diagrams for linear SEMs with correlated errors are ADMGs, and we may need exponential number of vanishing partial correlation tests based on the local Markov property given in [Richardson, 2003]. For this local test method to be applicable in models with correlated errors, it is therefore important to reduce the number of conditional independencies invoked by the local Markov property. It is known that normal distributions satisfy the composition axiom, which motivates our search for reduced local Markov property for probability distributions satisfying the composition axiom.

In Section 2, we give basic notation and definitions, and present the local Markov property developed in [Richardson, 2003]. In Section 3, we show that for a class of ADMGs, the local Markov property for probability distributions satisfying the composition axiom will invoke only one conditional independence relation for each vertex. In Section 4, we provide two lemmas under which the local Markov property can be reduced for a general ADMG, one assuming the composition axiom and the other not. We also provide a procedure that will incorporate the two lemmas and list all the conditional independence relations invoked by the reduced local Markov property. In Section 5, we show the usefulness of the results in Section 3 and 4 in testing linear SEMs. Section 6 concludes the paper.

## 2 Background

### 2.1 Notation and Definitions

For a vertex $x$ in an ADMG $G$, $\text{pa}_G(x) \equiv \{v|v \to x \text{ in } G\}$ is the set of *parents* of $x$. $\text{sp}_G(x) \equiv \{v|v \leftrightarrow x \text{ in } G\}$ is the set of *spouses* of $x$. $\text{an}_G(x) \equiv \{v|v \to \cdots \to x \text{ in } G \text{ or } v = x\}$ is the set of *ancestors* of $x$. And $\text{de}_G(x) \equiv \{v|v \leftarrow \cdots \leftarrow x \text{ in } G \text{ or } v = x\}$ is the set of *descendants* of $x$. These definitions will be applied to sets of vertices, so that, for example, $\text{pa}_G(A) \equiv \cup_{x \in A} \text{pa}_G(x)$, $\text{sp}_G(A) \equiv \cup_{x \in A} \text{sp}_G(x)$, etc.

A path is said to be a *mixed directed path from $\alpha$ to $\beta$* if it contains at least one directed edge and every edge on the path is either of the form $\gamma \leftrightarrow \delta$, or $\gamma \to \delta$ with $\delta$ between $\gamma$ and $\beta$. A mixed directed path from $\alpha$ to $\beta$ together with an edge $\beta \to \alpha$ or $\beta \leftrightarrow \alpha$ is called a *mixed directed cycle*.

For example, the path $a \to c \leftrightarrow d \to b \leftrightarrow a$ in the graph in Figure 1 forms a mixed directed cycle.

A non-endpoint vertex $z$ on a path is called a *collider* if two arrowheads on the path meet at $z$, i.e. $\to z \leftarrow$, $\leftrightarrow z \leftrightarrow$, $\leftrightarrow z \leftarrow$, $\to z \leftrightarrow$; all other non-endpoint vertices

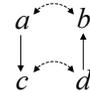

Figure 1: An ADMG with a mixed directed cycle.

on a path are *non-colliders*, i.e. $\leftarrow z \to$, $\leftarrow z \leftarrow$, $\to z \to$, $\leftrightarrow z \to$, $\leftarrow z \leftrightarrow$. A path between vertices $x$ and $y$ in an ADMG is said to be *m-connecting given a set* of vertices $Z$ if

(i) every non-collider on the path is not in $Z$, and
(ii) every collider on the path is an ancestor of a vertex in $Z$.

If there is no path m-connecting $x$ and $y$ given $Z$, then $x$ and $y$ are said to be *m-separated* given $Z$. Sets $X$ and $Y$ are said to be *m-separated* given $Z$, if for every pair $x, y$, with $x \in X$ and $y \in Y$, $x$ and $y$ are m-separated given $Z$.

A probability distribution $P$ is said to satisfy the *m-separation global Markov property* for $G$ if for arbitrary disjoint sets $X, Y, Z$,

$X$ is m-separated from $Y$ given $Z$ in $G \Rightarrow I(X, Z, Y)$

where $I(X, Z, Y)$ denotes that $X$ is conditionally independent of $Y$ given $Z$. The set of probability distributions that satisfy the m-separation global Markov property with respect to $G$ is denoted $P_m$.

It is well-known that probabilistic conditional independencies satisfy the following so-called *semi-graphoid axioms* [Pearl, 1988]:

- Symmetry
$$I(X, Z, Y) \Longleftrightarrow I(Y, Z, X)$$

- Decomposition
$$I(X, Z, Y \cup W) \Longrightarrow I(X, Z, Y) \ \& \ I(X, Z, W)$$

- Weak Union
$$I(X, Z, Y \cup W) \Longrightarrow I(X, Z \cup W, Y)$$

- Contraction
$$I(X, Z, Y) \ \& \ I(X, Z \cup Y, W) \Longrightarrow I(X, Z, Y \cup W)$$

where $X$, $Y$, $Z$, and $W$ are disjoint sets of variables. Some probability distributions, for example normal distributions, also satisfy the following composition axiom

- Composition
$$I(X, Z, Y) \ \& \ I(X, Z, W) \Longrightarrow I(X, Z, Y \cup W)$$

## 2.2 The Ordered Local Markov Property for ADMGs

In this section, we describe the local Markov property introduced by [Richardson, 2003]. An ordering($\prec$) on the vertices of $G$ is said to be consistent with $G$ if $x \prec y \Rightarrow y \notin \mathrm{an}(x)$. Given a consistent ordering $\prec$, let $\mathrm{pre}_{G,\prec}(x) \equiv \{v | v \prec x \text{ or } v = x\}$. A *c-component* of $G$ is a maximal set of vertices in $G$ such that any two vertices in the set are connected by a path on which every edge is of the form $\leftrightarrow$; a vertex that is not connected to any bi-directed edge forms a c-component by itself. For example, the graph in Figure 1 is composed of c-components $\{a,b\}$ and $\{c,d\}$. The *district* of $x$ in $G$ is the c-component of $G$ that includes $x$. Thus,

$$\mathrm{dis}_G(x) \equiv \{v | v \leftrightarrow \cdots \leftrightarrow x \text{ in } G \text{ or } v = x\}.$$

For example, in Figure 1, we have $\mathrm{dis}_G(a) = \{a,b\}$ and $\mathrm{dis}_G(d) = \{c,d\}$. A set $A$ is said to be *ancestral* if it is closed under the ancestor relation, i.e. if $\mathrm{an}_G(A) = A$. Let $G_A$ denote the induced subgraph of $G$ on the vertex set $A$, formed by removing from $G$ all vertices that are not in $A$, and all edges that do not have both endpoints in $A$. If $A$ is an ancestral set in an ADMG $G$, and $x$ is a vertex in $A$ that has no children in $A$ then the *Markov blanket of vertex $x$ with respect to the induced subgraph on $A$*, denoted $\mathrm{mb}(x,A)$ is defined to be

$$\mathrm{mb}(x,A) \equiv \mathrm{pa}_{G_A}(\mathrm{dis}_{G_A}(x)) \cup (\mathrm{dis}_{G_A}(x) \setminus \{x\}).$$

For example, for an ancestral set $A = \mathrm{an}_G(\{a,c\}) = \{a,c,d,e\}$ in Figure 2, we have

$$\mathrm{mb}(a,A) = \{d,c\}.$$

A probability distribution $P$ satisfies the *ordered local Markov property* for $G$ with respect to a consistent ordering $\prec$, if, for any $x$ and ancestral set $A$ such that $x \in A \subseteq \mathrm{pre}_{G,\prec}(x)$,

$$I(\{x\}, \mathrm{mb}(x,A), A \setminus (\mathrm{mb}(x,A) \cup \{x\})). \quad (1)$$

The set of probability distributions that satisfy the ordered local Markov property for $G$ under ordering $\prec$ is denoted $P_l(G, \prec)$.

The following theorem [Richardson, 2003] shows the equivalence between the ordered local Markov property and the global Markov property.

**Theorem 1** *[Richardson, 2003] If $G$ is an ADMG and $\prec$ is a consistent ordering then*

$$P_m(G) = P_l(G, \prec).$$

Therefore the (smaller) set of conditional independencies specified in the local Markov property (1) will imply all other conditional independencies which hold under the global Markov property. It is possible to further reduce the number of conditional independence relations in the local Markov property (1). An ancestral set $A$, with $x \in A \subseteq \mathrm{pre}_{G,\prec}(x)$ is said to be *maximal with respect to the Markov blanket* $\mathrm{mb}(x,A)$ if, whenever there is a set $B$ such that $A \subseteq B \subseteq \mathrm{pre}_{G,\prec}(x)$ and $\mathrm{mb}(x,A) = \mathrm{mb}(x,B)$, then $A = B$. For example, suppose that we are given an ordering $\prec: h \prec f \prec i \prec g \prec a \prec b \prec e \prec d \prec c$ for the graph $G$ in Figure 3(a). While an ancestral set $A = \mathrm{an}_G(\{a,c\}) = \{a,f,h,c,g,i,e\}$ is maximal with respect to the Markov blanket $\mathrm{mb}(c,A) = \{g,e\}$, an ancestral set $A' = \mathrm{an}_G(\{c\}) = \{i,g,e\}$ is not. It is shown that we only need to consider ancestral sets $A$ which are maximal with respect to $\mathrm{mb}(x,A)$ in the local Markov property (1) [Richardson, 2003].

Even though we only consider maximal ancestral sets, the ordered local Markov property may still invoke exponential number of conditional independence relations. For example, for a vertex $x$, if $\mathrm{dis}_G(x) \subseteq \mathrm{pre}_{G,\prec}(x)$ and $\mathrm{dis}_G(x)$ has a clique of $n$ vertices joined by bi-directed edges, then there are at least $O(2^{n-1})$ different Markov blankets.

## 3 ADMGs without Mixed Directed Cycles Assuming Composition Axiom

In this section, we show that if an ADMG has no mixed directed cycle and the probability distribution satisfies composition axiom, then a linear number of conditional independence relations are enough to imply all the other conditional independence relations which hold under the global Markov property.

Let $P_{m,comp}(G)$ be the set of probability distributions obeying composition axiom that satisfy the m-separation global Markov property with respect to $G$. Let $V$ be the set of vertices in $G$. Let

$$\mathrm{f}(x,G) \equiv \mathrm{pa}_G(x) \cup \mathrm{de}_G(\{x\} \cup \mathrm{sp}_G(x)). \quad (2)$$

Let $P_{l,comp}(G)$ be the set of probability distributions obeying composition axiom that satisfy the following local Markov property:

$$\forall x \in V, \quad I(\{x\}, \mathrm{pa}_G(x), V \setminus \mathrm{f}(x,G)). \quad (3)$$

**Theorem 2** *If an ADMG $G$ has no mixed directed cycle, then*

$$P_{l,comp}(G) = P_{m,comp}(G).$$

**Proof:** Let $P_{l,comp}(G, \prec)$ be the set of probability distributions that satisfy composition axiom and the ordered local Markov property for $G$ under ordering $\prec$. By Theorem 1, for any consistent ordering $\prec$, we have $P_{l,comp}(G, \prec) = P_{m,comp}(G)$. We show that for some consistent ordering $\prec$, $P_{l,comp}(G) \subseteq P_{l,comp}(G, \prec)$ and $P_{m,comp}(G) \subseteq P_{l,comp}(G)$.

To show $P_{m,comp}(G) \subseteq P_{l,comp}(G)$, we need to prove that any vertex $x$ is m-separated from $V \setminus \mathrm{f}(x, G)$ given $\mathrm{pa}_G(x)$ in $G$ with no mixed directed cycle. Suppose some vertex $x$ is not m-separated from $V \setminus \mathrm{f}(x, G)$ given $\mathrm{pa}_G(x)$. This would be true only if some vertex in $\mathrm{sp}_G(x)$ is an ancestor of $x$. Then, there would be a mixed directed cycle involving $x$.

We now show that we can construct a consistent ordering $\prec$ such that $P_{l,comp}(G) \subseteq P_{l,comp}(G, \prec)$ holds.

We do the following to get the desired ordering.

1. Given $G = (V, E)$, we combine all vertices in a c-component into one vertex. Let $V'$ be the resulting set of vertices. For each $x' \in V'$, let $\mathrm{cm}(x')$ be the set of corresponding vertices in $V$ which are combined into $x'$. We put a directed edge from $\alpha \in V'$ to $\beta \in V'$ if and only if there is a directed edge from some vertex in $\mathrm{cm}(\alpha)$ to some vertex in $\mathrm{cm}(\beta)$ (Since $G$ has no mixed directed cycle, if there is a directed edge from some vertex in $\mathrm{cm}(\alpha)$ to some vertex in $\mathrm{cm}(\beta)$ then there exists no directed edge from any vertex in $\mathrm{cm}(\beta)$ to any vertex in $\mathrm{cm}(\alpha)$). Let $G' = (V', E')$ be the resulting graph. Then $G'$ is a DAG because $G$ has no mixed directed cycle.

2. Let $\prec_{G'}$ be any consistent ordering on $V'$. Replace each vertex $x'$ in $\prec_{G'}$ with the set of vertices $\mathrm{cm}(x')$ arbitrarily ordered. Let $\prec_G$ be the resulting ordering.

Since the vertices in every c-component are consecutive in $\prec_G$, for any $x$ in $V$,

$$\mathrm{pre}_{G,\prec_G}(x) \cap (\mathrm{de}_G(\mathrm{dis}_G(x)) \setminus \mathrm{dis}_G(x)) = \emptyset. \quad (4)$$

Next, we show that $P_{l,comp}(G) \subseteq P_{l,comp}(G, \prec_G)$ holds. Let $P \in P_{l,comp}(G)$. We will show that the set of conditional independence relations in (3) imply the following conditional independence relations given by the ordered local Markov property for the vertex $x$.

For any maximal ancestral set $A$ such that

$$x \in A \subseteq \mathrm{pre}_{G,\prec_G}(x)$$
$$I(\{x\}, \mathrm{mb}(x, A), A \setminus (\mathrm{mb}(x, A) \cup \{x\})). \quad (5)$$

First, observe that for any vertex $y$ in $\mathrm{dis}_{G_A}(x)$, we have

$$A \setminus (\mathrm{pa}_{G_A}(y) \cup \{y\} \cup \mathrm{sp}_{G_A}(y)) \subseteq V \setminus \mathrm{f}(y, G), \quad (6)$$

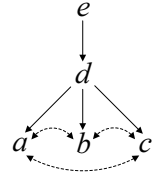

Figure 2: An ADMG with no mixed directed cycle.

since $A \subseteq V$ and $A \cap \mathrm{f}(y, G) = \mathrm{pa}_{G_A}(y) \cup \{y\} \cup \mathrm{sp}_{G_A}(y)$.
So, by (3), for all $y$ in $\mathrm{dis}_{G_A}(x)$, we have

$$I(\{y\}, \mathrm{pa}_{G_A}(y), A \setminus (\mathrm{pa}_{G_A}(y) \cup \{y\} \cup \mathrm{sp}_{G_A}(y))). \quad (7)$$

Let $S_1 = \mathrm{pa}_{G_A}(\mathrm{dis}_{G_A}(x)) \setminus \mathrm{pa}_{G_A}(y)$ and $S_2 = A \setminus (\mathrm{mb}(x, A) \cup \{x\})$.

It follows that

$$S_1 \subseteq A \setminus (\mathrm{pa}_{G_A}(y) \cup \{y\} \cup \mathrm{sp}_{G_A}(y)) \text{ and} \quad (8)$$
$$S_2 \subseteq A \setminus (\mathrm{pa}_{G_A}(y) \cup \{y\} \cup \mathrm{sp}_{G_A}(y)). \quad (9)$$

Also, we have

$$S_1 \cap S_2 = \emptyset, \quad (10)$$

since $S_1 \subseteq \mathrm{mb}(x, A)$. So,

$$I(\{y\}, \mathrm{pa}_{G_A}(y), S_1 \cup S_2) \text{ by decomposition} \quad (11)$$
$$I(\{y\}, \mathrm{pa}_{G_A}(y) \cup S_1, S_2) \text{ by weak union} \quad (12)$$
$$I(\mathrm{dis}_{G_A}(x), \mathrm{pa}_{G_A}(\mathrm{dis}_{G_A}(x)), \quad (13)$$
$$A \setminus (\mathrm{mb}(x, A) \cup \{x\}) \text{ by composition}$$
$$I(\{x\}, \mathrm{pa}_{G_A}(\mathrm{dis}_{G_A}(x)) \cup (\mathrm{dis}_{G_A}(x) \setminus \{x\}),$$
$$A \setminus (\mathrm{mb}(x, A) \cup \{x\})) \text{ by weak union.} \quad (14)$$

That is,

$$I(\{x\}, \mathrm{mb}(x, A), A \setminus (\mathrm{mb}(x, A) \cup \{x\})). \quad (15)$$

∎

As an example, consider the ADMG in Figure 2 which has no mixed directed cycles. For a consistent ordering $e \prec d \prec a \prec b \prec c$, the ordered local Markov property (1) for maximal ancestral sets $A$ involves the following conditional independencies

$$I(\{a\}, \{d\}, \{e\}), I(\{b\}, \{d\}, \{e\}), I(\{b\}, \{a, d\}, \{e\}),$$
$$I(\{c\}, \{d\}, \{e\}), I(\{c\}, \{a, d\}, \{e\}), I(\{c\}, \{b, d\}, \{e\}),$$
$$I(\{c\}, \{a, b, d\}, \{e\}). \quad (16)$$

The local Markov property in (3) invokes the following conditional independencies

$$I(\{a\}, \{d\}, \{e\}), I(\{b\}, \{d\}, \{e\}), I(\{c\}, \{d\}, \{e\}), \quad (17)$$

which, by Theorem 2 imply other conditional independencies in (16).

For the special case of graphs containing only bi-directed edges,[1] [Kauermann, 1996] provides a local Markov property for probability distributions obeying the composition axiom as follows:

$$\forall x \in V, \quad I(\{x\}, \emptyset, V \setminus (\{x\} \cup \mathrm{sp}_G(x))). \tag{18}$$

Since a graph containing only bi-directed edges is a special case of ADMGs without mixed directed cycles, the local Markov property given in (3) is applicable, and it turns out that (3) reduces to (18) for graphs containing only bi-directed edges. Therefore the local Markov property in (3) includes that given in [Kauermann, 1996] as a special case.

## 4  Reducing the Local Markov Property

When an ADMG $G$ has mixed directed cycles, the conditional independencies in (3) may not even hold in $G$. In this section, we show that we can still reduce the number of the conditional independence relations in the ordered local Markov property (1) with or without assuming the composition axiom

The following lemma allows us to remove some redundant conditional independence relations assuming composition axiom.

**Lemma 1** *Given a consistent ordering $\prec$, suppose for a vertex $x$, the following holds.*

$$\text{All vertices in } dis_G(x) \cap pre_{G,\prec}(x) \tag{19}$$
$$\text{are consecutive in } \prec$$

*and*

$$\text{for any two vertices } \alpha, \beta \text{ in } dis_G(x) \cap pre_{G,\prec}(x),$$
$$\text{there is no directed edge between } \alpha \text{ and } \beta. \tag{20}$$

*Then, any vertex $y$ in $dis_G(x) \cap pre_{G,\prec}(x)$ is m-separated from $V \setminus f(y, G)$ given $pa_G(y)$.*

*Assume that the probability distribution obeys the composition axiom. If, for all $y \in dis_G(x) \cap pre_{G,\prec}(x)$,*

$$I(\{y\}, pa_G(y), V \setminus f(y, G)), \tag{21}$$

*then*

*for any maximal ancestral set $A$ such that*
$$x \in A \subseteq pre_{G,\prec}(x)$$
$$I(\{x\}, mb(x, A), A \setminus (mb(x, A) \cup \{x\})). \tag{22}$$

---
[1][Kauermann, 1996] actually used undirected graphs with dashed edges which are Markov equivalent to graphs with only bi-directed edges (see [Richardson, 2003] for discussions).

**Proof:** Consider the first statement. Suppose some vertex $y$ in $\mathrm{dis}_G(x) \cap \mathrm{pre}_{G,\prec}(x)$ is not m-separated from $V \setminus \mathrm{f}(y, G)$ given $\mathrm{pa}_G(y)$. This would be possible only if a vertex in $\mathrm{sp}_G(y)$ were an ancestor of $y$. Suppose a vertex in $\mathrm{sp}_G(y)$ were a parent of $y$. Then, (20) would not hold. Also, if a vertex in $\mathrm{sp}_G(y)$ were an ancestor of but not parent of $y$, then (19) would not hold.

The proof of the second statement is the same as that of Theorem 2 except that (7) comes from (21) not (3).
∎

Although it is not possible to place all vertices in every c-component consecutively in a consistent ordering for a graph with mixed directed cycles, we can still put all vertices in some c-components consecutively or at least some subset of the vertices in some c-components consecutively in an ordering and then we can apply Lemma 1 for these c-components.

We now give a condition by which a conditional independence relation renders another conditional independence relation redundant.

**Lemma 2** *Given an ADMG $G$, a consistent ordering $\prec$ for $G$ and a vertex $x$, assume that for all $y \in pre_{G,\prec}(x) \setminus \{x\}$, and maximal ancestral set $S$ such that $y \in S \subseteq pre_{G,\prec}(y)$,*

$$I(\{y\}, mb(y, S), S \setminus (mb(y, S) \cup \{y\})). \tag{23}$$

*Let $A = pre_{G,\prec}(x)$ and $A'$ be a maximal ancestral set such that $x \in A' \subseteq pre_{G,\prec}(x)$. Let $Y = dis_{G_A}(x) \setminus dis_{G_{A'}}(x)$, $Y_1 = dis_{G_A}(x) \setminus (A' \cap dis_{G_A}(x))$ and $Y_2 = Y \setminus Y_1$. If $Y_2 = \emptyset$ and $pa_G(Y) \subseteq mb(x, A')$, then*

$$I(\{x\}, mb(x, A), A \setminus (mb(x, A) \cup \{x\})) \tag{24}$$

*implies*

$$I(\{x\}, mb(x, A'), A' \setminus (mb(x, A') \cup \{x\})). \tag{25}$$

**Proof:** We have

$$\mathrm{mb}(x, A) = \mathrm{mb}(x, A') \cup Y \cup \mathrm{pa}_G(Y) \tag{26}$$

and

$$A \setminus (\mathrm{mb}(x, A) \cup \{x\})$$
$$= \Big(A' \setminus (\mathrm{mb}(x, A') \cup \{x\} \cup Y_2 \cup \mathrm{pa}_G(Y))\Big)$$
$$\cup \Big(\mathrm{de}_G(Y_1) \setminus Y_1\Big) \tag{27}$$

Plugging (26) and (27) into (24), we get

$$I\Big(\{x\}, \mathrm{mb}(x, A') \cup Y \cup \mathrm{pa}_G(Y),$$
$$\Big(A' \setminus (\mathrm{mb}(x, A') \cup \{x\} \cup Y_2 \cup \mathrm{pa}_G(Y))\Big)$$
$$\cup \Big(\mathrm{de}_G(Y_1) \setminus Y_1\Big)\Big). \tag{28}$$

It follows from decomposition axiom that

$$I(\{x\}, \mathrm{mb}(x, A') \cup Y \cup \mathrm{pa}_G(Y),$$
$$A' \setminus (\mathrm{mb}(x, A') \cup \{x\} \cup Y_2 \cup \mathrm{pa}_G(Y))). \quad (29)$$

Also we have

$$I(Y_1, \mathrm{mb}(x, A') \cup Y_2 \cup \mathrm{pa}_G(Y),$$
$$A' \setminus (\mathrm{mb}(x, A') \cup \{x\} \cup Y_2 \cup \mathrm{pa}_G(Y))), \quad (30)$$

since $Y_1$ is m-separated from $A' \setminus (\mathrm{mb}(x, A') \cup \{x\} \cup Y_2 \cup \mathrm{pa}_G(Y))$ given $\mathrm{mb}(x, A') \cup Y_2 \cup \mathrm{pa}_G(Y)$ and by (23). Then, from (29), (30) and contraction axiom, we have

$$I(\{x\}, \mathrm{mb}(x, A') \cup Y_2 \cup \mathrm{pa}_G(Y),$$
$$A' \setminus (\mathrm{mb}(x, A') \cup \{x\} \cup Y_2 \cup \mathrm{pa}_G(Y))). \quad (31)$$

Since $Y_2 = \emptyset$ and $\mathrm{pa}_G(Y) \subseteq \mathrm{mb}(x, A')$,

$$I(\{x\}, \mathrm{mb}(x, A'), A' \setminus (\mathrm{mb}(x, A') \cup \{x\})) \quad (32)$$

holds. ∎

Note that Lemma 2 can be applied to probability distributions that do not satisfy composition axiom since the proof does not depend on composition axiom. Thus, this lemma can be used to reduce some redundant conditional independence relations for a general probability distribution.

Using the two lemmas above, we now give a procedure that, given a general ADMG, produces the set of conditional independence relations $R$ needed to derive the global Markov property for probability distributions obeying composition axiom. We do the following.

1. We generate a consistent ordering that will give as few conditional independence relations as possible. The method in Section 3 is modified a little bit to deal with mixed directed cycles. For each bi-directed edge $\alpha \leftrightarrow \beta$, check if there exists a mixed directed path between $\alpha$ and $\beta$.

   - If there is no mixed directed path between $\alpha$ and $\beta$, then combine the two vertices into one. If there is a vertex $\gamma$ which is adjacent to both $\alpha$ and $\beta$, then there are 9 possibilities for the edges among $\alpha, \beta,$ and $\gamma$. However, only 3 cases are relevant. (i) If we have $\alpha \to \gamma, \beta \to \gamma$, then these become one edge $\alpha\beta \to \gamma$. (ii) If we have $\alpha \leftarrow \gamma, \beta \leftarrow \gamma$, then these become one edge $\alpha\beta \leftarrow \gamma$. (iii) If we have $\alpha \leftrightarrow \gamma, \beta \leftrightarrow \gamma$, then these become one edge $\alpha\beta \leftrightarrow \gamma$. Other cases would imply that there is a mixed directed cycle involving the 3 vertices.

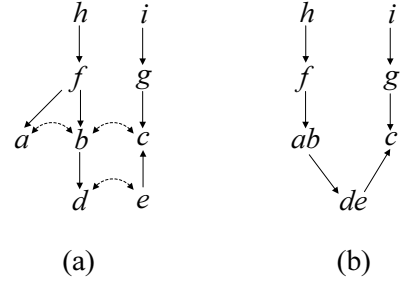

Figure 3: An example ADMG.

   - If there is a mixed directed path between $\alpha$ and $\beta$, then remove the bi-directed edge between $\alpha$ and $\beta$.

   We repeat this until no bi-directed edge is left. Let $G'$ be the resulting DAG. Then, we topologically sort $G'$ and get a consistent ordering $\prec_{G'}$ for $G'$. We replace each combined vertex $x'$ in $\prec_{G'}$ with the original vertices in $G$ arbitrarily ordered. Let $\prec_G$ be the resulting ordering.

2. Let $R = \emptyset$. We examine every vertex $x$ in $\prec_G$ starting from the first vertex in the ordering. If the conditions (19) and (20) in Lemma 1 for $x$ are satisfied, then do
   $R \leftarrow R \cup I(\{x\}, \mathrm{pa}_G(x), V \setminus \mathrm{f}(x, G))$.
   Otherwise, we do the following. Let $A = \mathrm{pre}_{G, \prec_G}(x)$. First, do
   $R \leftarrow R \cup I(\{x\}, \mathrm{mb}(x, A), A \setminus (\mathrm{mb}(x, A) \cup \{x\}))$.
   Then, we check for each smaller ancestral set $A'$ if $I(\{x\}, \mathrm{mb}(x, A), A \setminus (\mathrm{mb}(x, A) \cup \{x\}))$ implies $I(\{x\}, \mathrm{mb}(x, A'), A' \setminus (\mathrm{mb}(x, A') \cup \{x\}))$ by the condition described in Lemma 2. If the condition is not met, we do
   $R \leftarrow R \cup I(\{x\}, \mathrm{mb}(x, A'), A' \setminus (\mathrm{mb}(x, A') \cup \{x\}))$.
   We repeat this until every maximal ancestral set has been examined.

Note that, in the above Step 2, when we find that the conditions (19) and (20) in Lemma 1 are satisfied for a vertex $x$, only one conditional independence relation $I(\{x\}, \mathrm{pa}_G(x), V \setminus \mathrm{f}(x, G))$ needs to be added to $R$, since the conditional independence relations $I(\{y\}, \mathrm{pa}_G(y), V \setminus \mathrm{f}(y, G))$ for all $y$ in $(\mathrm{dis}_G(x) \cap \mathrm{pre}_{G, \prec}(x)) \setminus \{x\}$ must have been added to $R$ in the previous steps, and thus by Lemma 1 we can derive (22) from these conditional independence relations.

We show the application of the preceding procedure by considering the graph $G$ in Figure 3(a). In Step 1, $G$ is converted into $G'$ shown in Figure 3(b). Note that the bi-directed edge $b \leftrightarrow c$ is removed because there is a mixed directed path $b \to d \leftrightarrow e \to c$ between $b$ and

c. Let $\prec_{G'} = h \prec f \prec i \prec g \prec ab \prec ed \prec c$. Then, $\prec_G = h \prec f \prec i \prec g \prec a \prec b \prec e \prec d \prec c$. For $x = h, f, i, g, a, b, e$ and $d$, the condition in Lemma 1 is satisfied. For example, for $b$, we have

$$\text{dis}_G(b) \cap \text{pre}_{G, \prec_G}(b)$$
$$= \{a, b, c\} \cap \{h, f, i, g, a, b\} = \{a, b\}.$$

$a$ and $b$ are consecutive in $\prec_G$ and there is no directed edge between $a$ and $b$. So, only one conditional independence relation for each $x \in \{h, f, i, g, a, b, e, d\}$ is added to $R$. However, for $c$, the condition in Lemma 1 is not met since vertices in

$$\text{dis}_G(c) \cap \text{pre}_{G, \prec_G}(c)$$
$$= \{a, b, c\} \cap \{h, f, i, g, a, b, e, d, c\}$$
$$= \{a, b, c\}$$

are not placed consecutively in $\prec_G$. So, we resort to the ordered local Markov property. The maximal ancestral sets we need to consider are

$$\begin{aligned}
\text{an}_G(\{a, d, c\}) &= \{h, f, i, g, a, b, e, d, c\} \\
\text{an}_G(\{d, c\}) &= \{h, f, i, g, b, e, d, c\} \\
\text{an}_G(\{c, a\}) &= \{h, f, i, g, a, e, c\}
\end{aligned}$$

Their corresponding conditional independence relations are

$$I(\{c\}, \{a, b, g, e, f\}, \{h, i, d\}),$$
$$I(\{c\}, \{b, g, e, f\}, \{h, i, d\}),$$
$$I(\{c\}, \{g, e\}, \{h, f, i, a\}).$$

First, we take the largest ancestral set $\text{an}_G(\{a, d, c\})$ and put its corresponding conditional independence relation $I(\{c\}, \{a, b, g, e, f\}, \{h, i, d\})$ into $R$. Then, we proceed to check for the ancestral set $\text{an}_G(\{d, c\})$ whether its corresponding conditional independence relation $I(\{c\}, \{b, g, e, f\}, \{h, i, d\})$ is implied by $I(\{c\}, \{a, b, g, e, f\}, \{h, i, d\})$. Let $A = \text{an}_G(\{a, d, c\}) = \{h, f, i, g, a, b, e, d, c\}$ and $A' = \text{an}_G(\{d, c\}) = \{h, f, i, g, b, e, d, c\}$. Then, we have

$$\begin{aligned}
Y &= \text{dis}_{G_A}(c) \setminus \text{dis}_{G_{A'}}(c) \\
&= \{a, b, c\} \setminus \{b, c\} \\
&= \{a\} \\
Y_1 &= \text{dis}_{G_A}(c) \setminus (A' \cap \text{dis}_{G_A}(c)) = \{a\} \\
Y_2 &= Y \setminus Y_1 = \emptyset \text{ and} \\
\text{pa}_G(Y) &= \{f\} \subseteq \{b, g, e, f\} = \text{mb}(c, A').
\end{aligned}$$

Thus, by Lemma 2, $I(\{c\}, \{b, g, e, f\}, \{h, i, d\})$ is implied by $I(\{c\}, \{a, b, g, e, f\}, \{h, i, d\})$ and will not be added to $R$.

Finally we check the ancestral set $\text{an}_G(\{c, a\}) = \{h, f, i, g, a, e, c\}$ and its corresponding conditional independence relation $I(\{c\}, \{g, e\}, \{h, f, i, a\})$. Let $A = \text{an}_G(\{a, d, c\}) = \{h, f, i, g, a, b, e, d, c\}$ and $A' = \text{an}_G(\{c, a\}) = \{h, f, i, g, a, e, c\}$. Then, we have

$$\begin{aligned}
Y &= \text{dis}_{G_A}(c) \setminus \text{dis}_{G_{A'}}(c) \\
&= \{a, b, c\} \setminus \{c\} \\
&= \{a, b\} \\
Y_1 &= \text{dis}_{G_A}(c) \setminus (A' \cap \text{dis}_{G_A}(c)) \\
&= \{a, b, c\} \setminus \{a, c\} = \{b\} \\
Y_2 &= Y \setminus Y_1 = \{a, b\} \setminus \{b\} \neq \emptyset.
\end{aligned}$$

The condition in Lemma 2 is not met in this case. Thus, we put the conditional independence relation $I(\{c\}, \{g, e\}, \{h, f, i, a\})$ into $R$.

Therefore, for the graph $G$ in Figure 3(a), 10 conditional independence relations (one each for $h, f, i, g, a, b, e, d$ and two for $c$) are sufficient to derive all the other conditional independence relations that hold under the global Markov property. As a comparison, the ordered local Markov property (1) invokes 13 conditional independencies (one each for $h, f, i, g, a, e$, two each for $b$ and $d$, and three for $c$).

## 5 Applications in Testing Linear SEMs

A linear SEM over a set of random variables $V = \{v_1, \ldots, v_n\}$ is given by a set of structural equations of the form

$$v_j = \sum_{i<j} c_{ji} v_i + \epsilon_j, \quad j = 1, \ldots, n, \qquad (33)$$

where the summation is over the variables in $V$ judged to be immediate causes of $v_j$. $c_{ji}$ is called a *path coefficient*. $\epsilon_j$'s represent "error" terms and are assumed to have normal distribution. The model structure can be represented by an ADMG $G$, called a *path diagram*, as follows: the nodes of $G$ are the variables $v_1, \ldots, v_n$; there is a directed edge from $v_i$ to $v_j$ in $G$ if $v_i$ appears in the structural equation for $v_j$, that is, $c_{ji} \neq 0$; there is a bi-directed edge between $v_i$ and $v_j$ if the error terms $\epsilon_i$ and $\epsilon_j$ have non-zero correlation. For example, the graph in Figure 2 can serve as the path diagram for the following SEM,

$$\begin{aligned}
e &= \epsilon_e \\
d &= c_1 e + \epsilon_d \\
a &= c_2 d + \epsilon_a \qquad (34) \\
b &= c_3 d + \epsilon_b \\
c &= c_4 d + \epsilon_c \\
\text{Cov}(\epsilon_a, \epsilon_b) &\neq 0 \\
\text{Cov}(\epsilon_a, \epsilon_c) &\neq 0 \\
\text{Cov}(\epsilon_b, \epsilon_c) &\neq 0
\end{aligned}$$

In a linear SEM, conditional independence relations will correspond to zero partial correlations, that is, a partial correlation $\rho_{xy.Z}$ vanishes ($\rho_{xy.Z} = 0$ if and only if $x$ is m-separated from $y$ given $Z$ in the path diagram [Spirtes *et al.*, 1998, Koster, 1999]).

One important task in SEM applications is to test the model against data. One approach for this task is to test for the hypothesis $\rho_{xy.Z} = 0$ in the data if whenever $x$ is m-separated from $y$ given $Z$ in the path diagram of the model. We only need to test for the set of vanishing partial correlations specified by the local Markov property which will imply all vanishing partial correlations that hold under the global Markov property. In general, we may need exponential number of vanishing partial correlation tests based on the ordered local Markov property (1) given in [Richardson, 2003]. For this test method to be applicable in practice, it is therefore important to reduce the number of conditional independencies invoked by the local Markov property. It is known that normal distributions satisfy the composition axiom, therefore the results presented in Section 3 and 4 can be used to reduce the number of vanishing partial correlation tests. As an example, consider the SEM given in (34). If we use the ordered local Markov property (1), then we need to test for the vanishing of the following set of partial correlations (see(16)):

$$\{\rho_{ae.d}, \rho_{be.d}, \rho_{be.ad}, \rho_{ce.d}, \rho_{ce.ad}, \rho_{ce.bd}, \rho_{ce.abd}\}. \quad (35)$$

On the other hand, based on the result in Section 3, we only need to test for the vanishing of the following (see(17)):

$$\{\rho_{ae.d}, \rho_{be.d}, \rho_{ce.d}\}. \quad (36)$$

The number of tests needed is substantially reduced.

## 6 Conclusion

We show that the potentially exponential number of conditional independence relations invoked by the local Markov property in ADMGs may be reduced if the probability distributions satisfy the composition axiom. In ADMGs with no mixed directed cycles, only linear number of conditional independence relations are required. In ADMGs with mixed directed cycles, we give two conditions under which reduction is possible and we provide a procedure for doing the reduction. The results have important applications in testing linear SEMs.

#### Acknowledgements

The authors thank the anonymous reviewers for helpful comments. This research was partly supported by NSF grant IIS-0347846.